# Skin Cancer Recognition using Deep Residual Network

Brij Rokad
brij.rokad@gmail.com
Vellore Institute of Technology
Vellore, TN, India

Dr. Sureshkumar Nagarajan
sureshkumar.n@vit.ac.in
Vellore Institute of Technology
Vellore, TN, India

*Abstract* – **The advances in technology have enabled people to access internet from every part of the world. But to date, access to healthcare in remote areas is sparse. This proposed solution aims to bridge the gap between specialist doctors and patients. This prototype will be able to detect skin cancer from an image captured by the phone or any other camera. The network is deployed on cloud server-side processing for an even more accurate result. The Deep Residual learning model has been used for predicting the probability of cancer for server side The ResNet has three parametric layers. Each layer has Convolutional Neural Network, Batch Normalization, Maxpool and ReLU. Currently the model achieves an accuracy of 77% on the ISIC - 2017 challenge.**

*Keywords* – *skin cancer, melanoma, Residual neural network*

## I. Introduction

Skin melanoma is a threatening tumor developed in the skin layer. Scientists and researchers have widely research regarding skin cancer because of its significance and localization procedure. It is essential to detect skin cancer early so that treatment planning process can be done properly.

Skin cancer is a type of disease in which group of cells grow abnormally forming malignant melanoma invading the surrounding tissue. Skin cancer is the third most common and the major cause of non-accidental death in human among the ages of 20 – 39 [1]. Malignant Melanoma is one of the most dangerous cancers and can be fatal if untreated. They create changes in skin surface and color, however early discovery of these progressions can be cured for around 90% of the cases. In 2018, American Academy of dermatology has evaluated that there will be 91,270 new instances of melanoma in the United States and 9,320 deaths from skin cancer. Around 132,000 new instances of skin cancer are analyzed worldwide every year, as per the WHO (World Health Organization).

Digital dermoscopy is one of the most widely used non-invasive, cost effective imaging tool to identify melanomas in patients. It helps in prior diagnosis and allows doctors to give proper treatment that can improve the survival rate.

For this we have proposed an end to end solution that would require only an image as an input. We have followed Deep ResNet approach inspired by [10]. Our aim is to make it easy for people to communicate using the model. There are numerous people who use the ASL around the world. A vision-based approach to our solution attempts to reduce the requirement of human translators and increase dependency on the user's phone for translation.

## II. Related Work

Traditionally, skin cancers were diagnosed by analyzing images for nodule development in the body. Manual diagnosis is often laborious, time consuming and may cause inter observer variability. With large volume of medical database, this process of analyzing the images with the help of doctors is hardly reliable [3]. CAD systems help in analyzing various forms of cancer

[2] by aiding the doctors as a second opinion reader.

Skin cancers can be non – melanoma or melanoma. Non-melanoma skin cancers include: Basal cell carcinomas and Squamous Cell Carcinomas. Non-melanoma types are rarely lethal, on the other hand melanoma skin cancer are lethal. If melanoma is not treated early and removed in time, it can penetrate and grow deep into the skin, it can also spread to other parts of the body. Effects of a late diagnosis of skin cancer can be very significant in terms of personal health.

Melanomas are easily misdiagnosed at an early stage, because it can be confused with benign entities. The diagnosis of melanoma is difficult because of the variability in clinical appearance and an absence of pigmentation. Melanoma can mimic a scar or different tumors [4,5]. Computer Aided diagnosis system extensively began in early 1980s, significant research has been done in using CAD systems for medical image analysis.

In [6] By implementing four step methods, in the first step preprocessing on the images has been done to remove the noisy artifacts like skin hair and light reflection. Second and third step is patch extraction and CNN, a window is patched around the pixels and after that fed it to CNN. CNN analyze the patch from Local texture and General structure. These fully connected layers' output has been labelled as 0 or 1 in post processing.

While in [7] Lumix SZ1 camera has been used to capture 166 tumor images. These images after then transformed to generate 5000 images by tuning the contrast. Convolutional Neural Network VGG-16 was applied to train on those images. This train network creates feature vectors which in term is used to train Support Vector Machine (SVM) classifier. SVM predicts the location of tumor with the recognition rate of 89.5%.

With the advent of large datasets and powerful Graphical Processing Units (GPUs), Convolutional Neural Networks (CNNs) have aided in significant progress in computer vision. [8] won the ImageNet challenge in 2012 using a CNN processing on GPU. Medical image diagnosis has been one of the key drivers in the past for leaps in computer vision, the U-Net architecture by [9] which introduced skip connections had performed extremely well on medical images and small size datasets. Classifying images have been hard especially at cellular level, the one which are taken at 100x zoom. [10] Residual learning framework eases the training of deep neural networks. Residual Network has depth up to 152 layers which are 8 time deeper than VGG nets, with lower complexity and by the error margin of 3.57% on ImageNet test set.

Visualizing the layers in a CNN help in understanding the features that influence the algorithm in reaching on a decision. It helps in exploring the reasons behind the failure of the model. The Grad-Cam approach by [11] uses gradients of any class flowing through the final layer to produce a coarse map of highlighting the activations by the model.

[12] has used a similar approach to achieve dermatologist level accuracy on skin cancer detection. Their method has used an Inception V3 [13] model pre-trained on the ImageNet. They used transfer learning with a loss across many granular class that inherit from a parent class namely melanoma, nevus and seborrheic keratosis.

Automated recognition of melanoma is a growing area of research. [14] It can be done with semi supervised, self-advised learning model. Data is categorized in form of labeled and unlabeled data set after that exponential function is applied to tune the data to maximize the separation of labeled data. Self-advise SVM is used to enhance the classification Deep network is trained using bootstrap technique to achieve 89% accuracy.

## III. Methodology

The main aim of this implementation is to detect skin cancer through RGB images, to achieve this, we build a deep learning model that is capable of extracting features from the given dataset. Also, to give a prediction on the go we build a separate skin cancer module that can run on portable devices.

Figure 1 shows the proposed architecture which we have used for melanoma recognition.

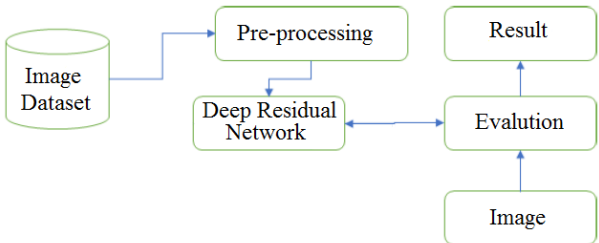

Fig. 1 Proposed Architecture

For translating the image to the relevant cancer recognition, we have trained the pre-trained model. Our model is trained on the Residual Network [10]. The trained model is then used for inference from the images that are being fed as input to the image. In accordance to the pre-trained model we process the image by dividing every pixel by 255 and then resizing the image to 244X244.

### A. Dataset

To predict the outcome a relevant dataset is needed to train the models. The dataset which is used here was taken from ISIC archive. For this system, we have use the ISIC - 2017 dataset [15]. It contains 2000 images. 374 - Melanoma, 254 - Seboherric Keratosis, 1372 - Nevus (Begnin).

### B. Image Pre-processing

Images from the dataset are pre-processed in order to get most efficient output. Here we have resized the images to 224x224 pixel, after that rotate, flip horizontally and vertically to train the network and subtract the mean value of image from every RGB channel of the image.

### C. Deep Residual Neural Network

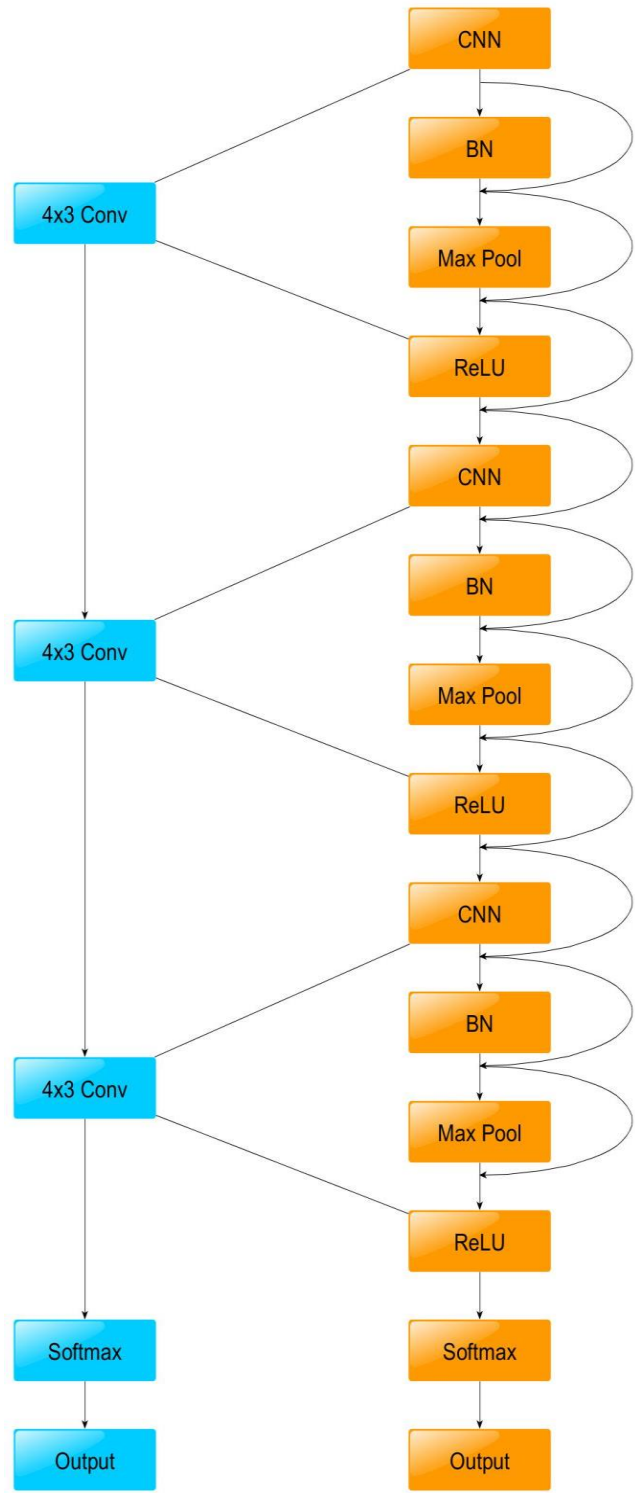

Fig. 2 Network architectures for ImageNet. A residual network with 3 parameter layers

In figure 2 right side of the parameter layers are Residual Network's architecture.

A parameter layer in ResNet has four sub layers. First layer is Convolutional Neural Network (CNN), it is responsible for detecting the shape and size of affected area. Second layer is Batch normalization (BN). BN is responsible for normalizing the image quality to get most efficient weights.

Third layer is Max pool layer. It works to maintain the image scale values by taking mean of the image RGB values and subtract it from every RGB value of the image, and the last layer is Rectified Linear Unit (ReLU). It acts as an activation function and also to update weights and bias.

After each parameter layer computation, the output will be send to next parameter layer with the initial inputs and the process will continue till last parameter layer.

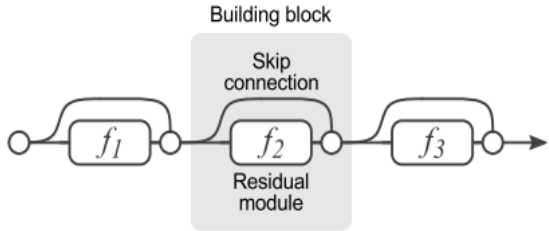

(a) Conventional 3-block residual network

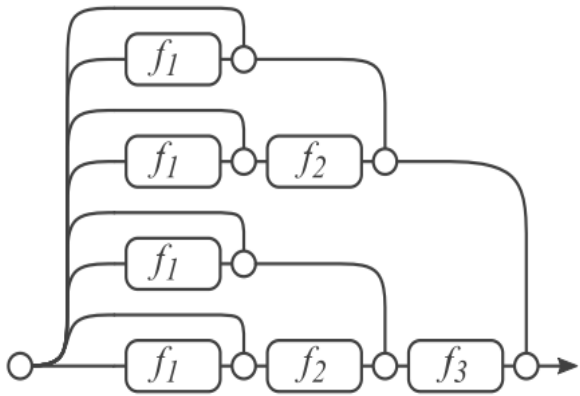

(b) Unraveled view of (a)

Fig. 3 ResNet

As shown in figure 3 every parameter layer can receive computed inputs from every previous parameter layer and also initial inputs.

Because of several inputs in each layer, the network can predict the outcome of the images more efficiently.

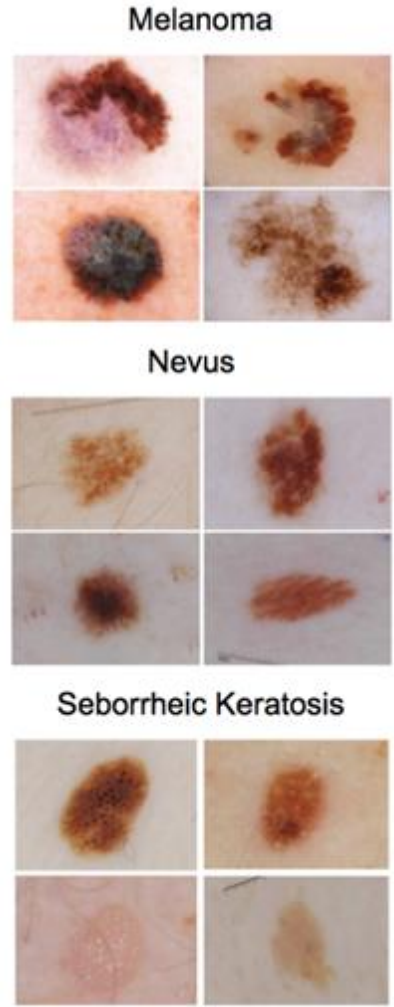

Fig. 4 ResNet classification

The ResNet has three classes of images as shown in figure 4 above. We have converted these three-class into a binary classification. By doing this the model has shown to improve its accuracy.

## IV. Results

The results below in figure 5 are of the model trained using a Residual Network visualized through the gradients-cam. From the image, it can be noticed that the deep learning model is activating precisely for the cancer instead of other features.

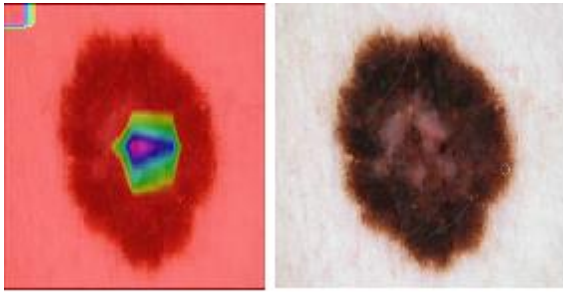
Fig. 5 Melanoma affected area

However, In the image below figure 6, the model is focusing on the complexion of the skin. There is scope for improvement of the model through better data augmentation and a faster inference is also needed.

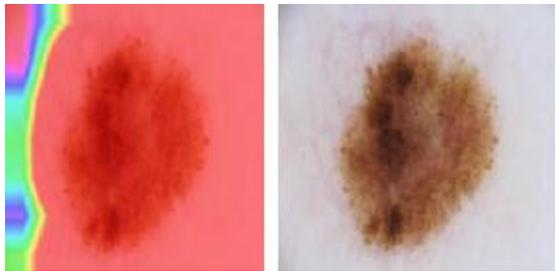
Fig. 6 Normal skin area

With training accuracy and validation accuracy there two more categories which are loss in training and loss in validation. The entire network relies on minimizing cross-entropy and the loss is calculated on training and validation as how well the model is doing. So, lower the loss, better the model. Loss is a summation of the errors made for each example in training or validation sets. On figure 7 we can clearly see that the loss at last epoch is minimum and validation accuracy is 77%

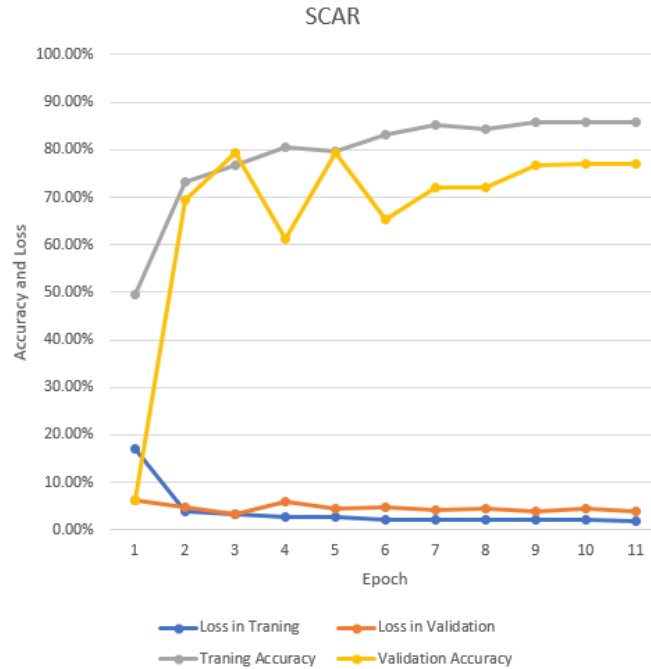
Fig. 7 Model Training Accuracy

## V. Conclusion

From the tests a maximum accuracy of 77% is attained. The network learned the characteristic of melanoma that allowed it to predict the skin cancer in real time. The model was developed with a light-weight architecture which enabled the complete architecture to be stored on a portable device. This helped with the accessibility of such a solution for the public. Hence, algorithmic recognition for portable devices is currently preferred in order to present the majority of people with a highly accessible solution. In the future, the dataset preprocessing will help to improve the accuracy of the model. The lighting conditions and distance of the image from the camera should not affect the outcome of the prediction.